\documentclass[letterpaper, 10 pt, conference]{ieeeconf}  

\IEEEoverridecommandlockouts                              

\usepackage{amsmath} 

\usepackage{graphicx}
\usepackage{caption}
\usepackage{subcaption}
\usepackage{tabularx}
\usepackage{bm}
\usepackage{amsfonts}
\usepackage{color}

\usepackage[draft]{minted}
\usepackage{tcolorbox}

\usepackage{mathtools}
\usepackage{arydshln}

\usepackage[symbol]{footmisc}

\makeatletter
\def\PYGdefault@reset{\let\PYGdefault@it=\relax \let\PYGdefault@bf=\relax%
    \let\PYGdefault@ul=\relax \let\PYGdefault@tc=\relax%
    \let\PYGdefault@bc=\relax \let\PYGdefault@ff=\relax}
\def\PYGdefault@tok#1{\csname PYGdefault@tok@#1\endcsname}
\def\PYGdefault@toks#1+{\ifx\relax#1\empty\else%
    \PYGdefault@tok{#1}\expandafter\PYGdefault@toks\fi}
\def\PYGdefault@do#1{\PYGdefault@bc{\PYGdefault@tc{\PYGdefault@ul{%
    \PYGdefault@it{\PYGdefault@bf{\PYGdefault@ff{#1}}}}}}}
\def\PYGdefault#1#2{\PYGdefault@reset\PYGdefault@toks#1+\relax+\PYGdefault@do{#2}}

\expandafter\def\csname PYGdefault@tok@gd\endcsname{\def\PYGdefault@tc##1{\textcolor[rgb]{0.63,0.00,0.00}{##1}}}
\expandafter\def\csname PYGdefault@tok@gu\endcsname{\let\PYGdefault@bf=\textbf\def\PYGdefault@tc##1{\textcolor[rgb]{0.50,0.00,0.50}{##1}}}
\expandafter\def\csname PYGdefault@tok@gt\endcsname{\def\PYGdefault@tc##1{\textcolor[rgb]{0.00,0.27,0.87}{##1}}}
\expandafter\def\csname PYGdefault@tok@gs\endcsname{\let\PYGdefault@bf=\textbf}
\expandafter\def\csname PYGdefault@tok@gr\endcsname{\def\PYGdefault@tc##1{\textcolor[rgb]{1.00,0.00,0.00}{##1}}}
\expandafter\def\csname PYGdefault@tok@cm\endcsname{\let\PYGdefault@it=\textit\def\PYGdefault@tc##1{\textcolor[rgb]{0.25,0.50,0.50}{##1}}}
\expandafter\def\csname PYGdefault@tok@vg\endcsname{\def\PYGdefault@tc##1{\textcolor[rgb]{0.10,0.09,0.49}{##1}}}
\expandafter\def\csname PYGdefault@tok@vi\endcsname{\def\PYGdefault@tc##1{\textcolor[rgb]{0.10,0.09,0.49}{##1}}}
\expandafter\def\csname PYGdefault@tok@mh\endcsname{\def\PYGdefault@tc##1{\textcolor[rgb]{0.40,0.40,0.40}{##1}}}
\expandafter\def\csname PYGdefault@tok@cs\endcsname{\let\PYGdefault@it=\textit\def\PYGdefault@tc##1{\textcolor[rgb]{0.25,0.50,0.50}{##1}}}
\expandafter\def\csname PYGdefault@tok@ge\endcsname{\let\PYGdefault@it=\textit}
\expandafter\def\csname PYGdefault@tok@vc\endcsname{\def\PYGdefault@tc##1{\textcolor[rgb]{0.10,0.09,0.49}{##1}}}
\expandafter\def\csname PYGdefault@tok@il\endcsname{\def\PYGdefault@tc##1{\textcolor[rgb]{0.40,0.40,0.40}{##1}}}
\expandafter\def\csname PYGdefault@tok@go\endcsname{\def\PYGdefault@tc##1{\textcolor[rgb]{0.53,0.53,0.53}{##1}}}
\expandafter\def\csname PYGdefault@tok@cp\endcsname{\def\PYGdefault@tc##1{\textcolor[rgb]{0.74,0.48,0.00}{##1}}}
\expandafter\def\csname PYGdefault@tok@gi\endcsname{\def\PYGdefault@tc##1{\textcolor[rgb]{0.00,0.63,0.00}{##1}}}
\expandafter\def\csname PYGdefault@tok@gh\endcsname{\let\PYGdefault@bf=\textbf\def\PYGdefault@tc##1{\textcolor[rgb]{0.00,0.00,0.50}{##1}}}
\expandafter\def\csname PYGdefault@tok@ni\endcsname{\let\PYGdefault@bf=\textbf\def\PYGdefault@tc##1{\textcolor[rgb]{0.60,0.60,0.60}{##1}}}
\expandafter\def\csname PYGdefault@tok@nl\endcsname{\def\PYGdefault@tc##1{\textcolor[rgb]{0.63,0.63,0.00}{##1}}}
\expandafter\def\csname PYGdefault@tok@nn\endcsname{\let\PYGdefault@bf=\textbf\def\PYGdefault@tc##1{\textcolor[rgb]{0.00,0.00,1.00}{##1}}}
\expandafter\def\csname PYGdefault@tok@no\endcsname{\def\PYGdefault@tc##1{\textcolor[rgb]{0.53,0.00,0.00}{##1}}}
\expandafter\def\csname PYGdefault@tok@na\endcsname{\def\PYGdefault@tc##1{\textcolor[rgb]{0.49,0.56,0.16}{##1}}}
\expandafter\def\csname PYGdefault@tok@nb\endcsname{\def\PYGdefault@tc##1{\textcolor[rgb]{0.00,0.50,0.00}{##1}}}
\expandafter\def\csname PYGdefault@tok@nc\endcsname{\let\PYGdefault@bf=\textbf\def\PYGdefault@tc##1{\textcolor[rgb]{0.00,0.00,1.00}{##1}}}
\expandafter\def\csname PYGdefault@tok@nd\endcsname{\def\PYGdefault@tc##1{\textcolor[rgb]{0.67,0.13,1.00}{##1}}}
\expandafter\def\csname PYGdefault@tok@ne\endcsname{\let\PYGdefault@bf=\textbf\def\PYGdefault@tc##1{\textcolor[rgb]{0.82,0.25,0.23}{##1}}}
\expandafter\def\csname PYGdefault@tok@nf\endcsname{\def\PYGdefault@tc##1{\textcolor[rgb]{0.00,0.00,1.00}{##1}}}
\expandafter\def\csname PYGdefault@tok@si\endcsname{\let\PYGdefault@bf=\textbf\def\PYGdefault@tc##1{\textcolor[rgb]{0.73,0.40,0.53}{##1}}}
\expandafter\def\csname PYGdefault@tok@s2\endcsname{\def\PYGdefault@tc##1{\textcolor[rgb]{0.73,0.13,0.13}{##1}}}
\expandafter\def\csname PYGdefault@tok@nt\endcsname{\let\PYGdefault@bf=\textbf\def\PYGdefault@tc##1{\textcolor[rgb]{0.00,0.50,0.00}{##1}}}
\expandafter\def\csname PYGdefault@tok@nv\endcsname{\def\PYGdefault@tc##1{\textcolor[rgb]{0.10,0.09,0.49}{##1}}}
\expandafter\def\csname PYGdefault@tok@s1\endcsname{\def\PYGdefault@tc##1{\textcolor[rgb]{0.73,0.13,0.13}{##1}}}
\expandafter\def\csname PYGdefault@tok@ch\endcsname{\let\PYGdefault@it=\textit\def\PYGdefault@tc##1{\textcolor[rgb]{0.25,0.50,0.50}{##1}}}
\expandafter\def\csname PYGdefault@tok@m\endcsname{\def\PYGdefault@tc##1{\textcolor[rgb]{0.40,0.40,0.40}{##1}}}
\expandafter\def\csname PYGdefault@tok@gp\endcsname{\let\PYGdefault@bf=\textbf\def\PYGdefault@tc##1{\textcolor[rgb]{0.00,0.00,0.50}{##1}}}
\expandafter\def\csname PYGdefault@tok@sh\endcsname{\def\PYGdefault@tc##1{\textcolor[rgb]{0.73,0.13,0.13}{##1}}}
\expandafter\def\csname PYGdefault@tok@ow\endcsname{\let\PYGdefault@bf=\textbf\def\PYGdefault@tc##1{\textcolor[rgb]{0.67,0.13,1.00}{##1}}}
\expandafter\def\csname PYGdefault@tok@sx\endcsname{\def\PYGdefault@tc##1{\textcolor[rgb]{0.00,0.50,0.00}{##1}}}
\expandafter\def\csname PYGdefault@tok@bp\endcsname{\def\PYGdefault@tc##1{\textcolor[rgb]{0.00,0.50,0.00}{##1}}}
\expandafter\def\csname PYGdefault@tok@c1\endcsname{\let\PYGdefault@it=\textit\def\PYGdefault@tc##1{\textcolor[rgb]{0.25,0.50,0.50}{##1}}}
\expandafter\def\csname PYGdefault@tok@o\endcsname{\def\PYGdefault@tc##1{\textcolor[rgb]{0.40,0.40,0.40}{##1}}}
\expandafter\def\csname PYGdefault@tok@kc\endcsname{\let\PYGdefault@bf=\textbf\def\PYGdefault@tc##1{\textcolor[rgb]{0.00,0.50,0.00}{##1}}}
\expandafter\def\csname PYGdefault@tok@c\endcsname{\let\PYGdefault@it=\textit\def\PYGdefault@tc##1{\textcolor[rgb]{0.25,0.50,0.50}{##1}}}
\expandafter\def\csname PYGdefault@tok@mf\endcsname{\def\PYGdefault@tc##1{\textcolor[rgb]{0.40,0.40,0.40}{##1}}}
\expandafter\def\csname PYGdefault@tok@err\endcsname{\def\PYGdefault@bc##1{\setlength{\fboxsep}{0pt}\fcolorbox[rgb]{1.00,0.00,0.00}{1,1,1}{\strut ##1}}}
\expandafter\def\csname PYGdefault@tok@mb\endcsname{\def\PYGdefault@tc##1{\textcolor[rgb]{0.40,0.40,0.40}{##1}}}
\expandafter\def\csname PYGdefault@tok@ss\endcsname{\def\PYGdefault@tc##1{\textcolor[rgb]{0.10,0.09,0.49}{##1}}}
\expandafter\def\csname PYGdefault@tok@sr\endcsname{\def\PYGdefault@tc##1{\textcolor[rgb]{0.73,0.40,0.53}{##1}}}
\expandafter\def\csname PYGdefault@tok@mo\endcsname{\def\PYGdefault@tc##1{\textcolor[rgb]{0.40,0.40,0.40}{##1}}}
\expandafter\def\csname PYGdefault@tok@kd\endcsname{\let\PYGdefault@bf=\textbf\def\PYGdefault@tc##1{\textcolor[rgb]{0.00,0.50,0.00}{##1}}}
\expandafter\def\csname PYGdefault@tok@mi\endcsname{\def\PYGdefault@tc##1{\textcolor[rgb]{0.40,0.40,0.40}{##1}}}
\expandafter\def\csname PYGdefault@tok@kn\endcsname{\let\PYGdefault@bf=\textbf\def\PYGdefault@tc##1{\textcolor[rgb]{0.00,0.50,0.00}{##1}}}
\expandafter\def\csname PYGdefault@tok@cpf\endcsname{\let\PYGdefault@it=\textit\def\PYGdefault@tc##1{\textcolor[rgb]{0.25,0.50,0.50}{##1}}}
\expandafter\def\csname PYGdefault@tok@kr\endcsname{\let\PYGdefault@bf=\textbf\def\PYGdefault@tc##1{\textcolor[rgb]{0.00,0.50,0.00}{##1}}}
\expandafter\def\csname PYGdefault@tok@s\endcsname{\def\PYGdefault@tc##1{\textcolor[rgb]{0.73,0.13,0.13}{##1}}}
\expandafter\def\csname PYGdefault@tok@kp\endcsname{\def\PYGdefault@tc##1{\textcolor[rgb]{0.00,0.50,0.00}{##1}}}
\expandafter\def\csname PYGdefault@tok@w\endcsname{\def\PYGdefault@tc##1{\textcolor[rgb]{0.73,0.73,0.73}{##1}}}
\expandafter\def\csname PYGdefault@tok@kt\endcsname{\def\PYGdefault@tc##1{\textcolor[rgb]{0.69,0.00,0.25}{##1}}}
\expandafter\def\csname PYGdefault@tok@sc\endcsname{\def\PYGdefault@tc##1{\textcolor[rgb]{0.73,0.13,0.13}{##1}}}
\expandafter\def\csname PYGdefault@tok@sb\endcsname{\def\PYGdefault@tc##1{\textcolor[rgb]{0.73,0.13,0.13}{##1}}}
\expandafter\def\csname PYGdefault@tok@k\endcsname{\let\PYGdefault@bf=\textbf\def\PYGdefault@tc##1{\textcolor[rgb]{0.00,0.50,0.00}{##1}}}
\expandafter\def\csname PYGdefault@tok@se\endcsname{\let\PYGdefault@bf=\textbf\def\PYGdefault@tc##1{\textcolor[rgb]{0.73,0.40,0.13}{##1}}}
\expandafter\def\csname PYGdefault@tok@sd\endcsname{\let\PYGdefault@it=\textit\def\PYGdefault@tc##1{\textcolor[rgb]{0.73,0.13,0.13}{##1}}}

\makeatother

\makeatletter
\def\PYG@reset{\let\PYG@it=\relax \let\PYG@bf=\relax%
    \let\PYG@ul=\relax \let\PYG@tc=\relax%
    \let\PYG@bc=\relax \let\PYG@ff=\relax}
\def\PYG@tok#1{\csname PYG@tok@#1\endcsname}
\def\PYG@toks#1+{\ifx\relax#1\empty\else%
    \PYG@tok{#1}\expandafter\PYG@toks\fi}
\def\PYG@do#1{\PYG@bc{\PYG@tc{\PYG@ul{%
    \PYG@it{\PYG@bf{\PYG@ff{#1}}}}}}}
\def\PYG#1#2{\PYG@reset\PYG@toks#1+\relax+\PYG@do{#2}}

\expandafter\def\csname PYG@tok@gd\endcsname{\def\PYG@tc##1{\textcolor[rgb]{0.63,0.00,0.00}{##1}}}
\expandafter\def\csname PYG@tok@gu\endcsname{\let\PYG@bf=\textbf\def\PYG@tc##1{\textcolor[rgb]{0.50,0.00,0.50}{##1}}}
\expandafter\def\csname PYG@tok@gt\endcsname{\def\PYG@tc##1{\textcolor[rgb]{0.00,0.27,0.87}{##1}}}
\expandafter\def\csname PYG@tok@gs\endcsname{\let\PYG@bf=\textbf}
\expandafter\def\csname PYG@tok@gr\endcsname{\def\PYG@tc##1{\textcolor[rgb]{1.00,0.00,0.00}{##1}}}
\expandafter\def\csname PYG@tok@cm\endcsname{\let\PYG@it=\textit\def\PYG@tc##1{\textcolor[rgb]{0.25,0.50,0.50}{##1}}}
\expandafter\def\csname PYG@tok@vg\endcsname{\def\PYG@tc##1{\textcolor[rgb]{0.10,0.09,0.49}{##1}}}
\expandafter\def\csname PYG@tok@vi\endcsname{\def\PYG@tc##1{\textcolor[rgb]{0.10,0.09,0.49}{##1}}}
\expandafter\def\csname PYG@tok@mh\endcsname{\def\PYG@tc##1{\textcolor[rgb]{0.40,0.40,0.40}{##1}}}
\expandafter\def\csname PYG@tok@cs\endcsname{\let\PYG@it=\textit\def\PYG@tc##1{\textcolor[rgb]{0.25,0.50,0.50}{##1}}}
\expandafter\def\csname PYG@tok@ge\endcsname{\let\PYG@it=\textit}
\expandafter\def\csname PYG@tok@vc\endcsname{\def\PYG@tc##1{\textcolor[rgb]{0.10,0.09,0.49}{##1}}}
\expandafter\def\csname PYG@tok@il\endcsname{\def\PYG@tc##1{\textcolor[rgb]{0.40,0.40,0.40}{##1}}}
\expandafter\def\csname PYG@tok@go\endcsname{\def\PYG@tc##1{\textcolor[rgb]{0.53,0.53,0.53}{##1}}}
\expandafter\def\csname PYG@tok@cp\endcsname{\def\PYG@tc##1{\textcolor[rgb]{0.74,0.48,0.00}{##1}}}
\expandafter\def\csname PYG@tok@gi\endcsname{\def\PYG@tc##1{\textcolor[rgb]{0.00,0.63,0.00}{##1}}}
\expandafter\def\csname PYG@tok@gh\endcsname{\let\PYG@bf=\textbf\def\PYG@tc##1{\textcolor[rgb]{0.00,0.00,0.50}{##1}}}
\expandafter\def\csname PYG@tok@ni\endcsname{\let\PYG@bf=\textbf\def\PYG@tc##1{\textcolor[rgb]{0.60,0.60,0.60}{##1}}}
\expandafter\def\csname PYG@tok@nl\endcsname{\def\PYG@tc##1{\textcolor[rgb]{0.63,0.63,0.00}{##1}}}
\expandafter\def\csname PYG@tok@nn\endcsname{\let\PYG@bf=\textbf\def\PYG@tc##1{\textcolor[rgb]{0.00,0.00,1.00}{##1}}}
\expandafter\def\csname PYG@tok@no\endcsname{\def\PYG@tc##1{\textcolor[rgb]{0.53,0.00,0.00}{##1}}}
\expandafter\def\csname PYG@tok@na\endcsname{\def\PYG@tc##1{\textcolor[rgb]{0.49,0.56,0.16}{##1}}}
\expandafter\def\csname PYG@tok@nb\endcsname{\def\PYG@tc##1{\textcolor[rgb]{0.00,0.50,0.00}{##1}}}
\expandafter\def\csname PYG@tok@nc\endcsname{\let\PYG@bf=\textbf\def\PYG@tc##1{\textcolor[rgb]{0.00,0.00,1.00}{##1}}}
\expandafter\def\csname PYG@tok@nd\endcsname{\def\PYG@tc##1{\textcolor[rgb]{0.67,0.13,1.00}{##1}}}
\expandafter\def\csname PYG@tok@ne\endcsname{\let\PYG@bf=\textbf\def\PYG@tc##1{\textcolor[rgb]{0.82,0.25,0.23}{##1}}}
\expandafter\def\csname PYG@tok@nf\endcsname{\def\PYG@tc##1{\textcolor[rgb]{0.00,0.00,1.00}{##1}}}
\expandafter\def\csname PYG@tok@si\endcsname{\let\PYG@bf=\textbf\def\PYG@tc##1{\textcolor[rgb]{0.73,0.40,0.53}{##1}}}
\expandafter\def\csname PYG@tok@s2\endcsname{\def\PYG@tc##1{\textcolor[rgb]{0.73,0.13,0.13}{##1}}}
\expandafter\def\csname PYG@tok@nt\endcsname{\let\PYG@bf=\textbf\def\PYG@tc##1{\textcolor[rgb]{0.00,0.50,0.00}{##1}}}
\expandafter\def\csname PYG@tok@nv\endcsname{\def\PYG@tc##1{\textcolor[rgb]{0.10,0.09,0.49}{##1}}}
\expandafter\def\csname PYG@tok@s1\endcsname{\def\PYG@tc##1{\textcolor[rgb]{0.73,0.13,0.13}{##1}}}
\expandafter\def\csname PYG@tok@ch\endcsname{\let\PYG@it=\textit\def\PYG@tc##1{\textcolor[rgb]{0.25,0.50,0.50}{##1}}}
\expandafter\def\csname PYG@tok@m\endcsname{\def\PYG@tc##1{\textcolor[rgb]{0.40,0.40,0.40}{##1}}}
\expandafter\def\csname PYG@tok@gp\endcsname{\let\PYG@bf=\textbf\def\PYG@tc##1{\textcolor[rgb]{0.00,0.00,0.50}{##1}}}
\expandafter\def\csname PYG@tok@sh\endcsname{\def\PYG@tc##1{\textcolor[rgb]{0.73,0.13,0.13}{##1}}}
\expandafter\def\csname PYG@tok@ow\endcsname{\let\PYG@bf=\textbf\def\PYG@tc##1{\textcolor[rgb]{0.67,0.13,1.00}{##1}}}
\expandafter\def\csname PYG@tok@sx\endcsname{\def\PYG@tc##1{\textcolor[rgb]{0.00,0.50,0.00}{##1}}}
\expandafter\def\csname PYG@tok@bp\endcsname{\def\PYG@tc##1{\textcolor[rgb]{0.00,0.50,0.00}{##1}}}
\expandafter\def\csname PYG@tok@c1\endcsname{\let\PYG@it=\textit\def\PYG@tc##1{\textcolor[rgb]{0.25,0.50,0.50}{##1}}}
\expandafter\def\csname PYG@tok@o\endcsname{\def\PYG@tc##1{\textcolor[rgb]{0.40,0.40,0.40}{##1}}}
\expandafter\def\csname PYG@tok@kc\endcsname{\let\PYG@bf=\textbf\def\PYG@tc##1{\textcolor[rgb]{0.00,0.50,0.00}{##1}}}
\expandafter\def\csname PYG@tok@c\endcsname{\let\PYG@it=\textit\def\PYG@tc##1{\textcolor[rgb]{0.25,0.50,0.50}{##1}}}
\expandafter\def\csname PYG@tok@mf\endcsname{\def\PYG@tc##1{\textcolor[rgb]{0.40,0.40,0.40}{##1}}}
\expandafter\def\csname PYG@tok@err\endcsname{\def\PYG@bc##1{\setlength{\fboxsep}{0pt}\fcolorbox[rgb]{1.00,0.00,0.00}{1,1,1}{\strut ##1}}}
\expandafter\def\csname PYG@tok@mb\endcsname{\def\PYG@tc##1{\textcolor[rgb]{0.40,0.40,0.40}{##1}}}
\expandafter\def\csname PYG@tok@ss\endcsname{\def\PYG@tc##1{\textcolor[rgb]{0.10,0.09,0.49}{##1}}}
\expandafter\def\csname PYG@tok@sr\endcsname{\def\PYG@tc##1{\textcolor[rgb]{0.73,0.40,0.53}{##1}}}
\expandafter\def\csname PYG@tok@mo\endcsname{\def\PYG@tc##1{\textcolor[rgb]{0.40,0.40,0.40}{##1}}}
\expandafter\def\csname PYG@tok@kd\endcsname{\let\PYG@bf=\textbf\def\PYG@tc##1{\textcolor[rgb]{0.00,0.50,0.00}{##1}}}
\expandafter\def\csname PYG@tok@mi\endcsname{\def\PYG@tc##1{\textcolor[rgb]{0.40,0.40,0.40}{##1}}}
\expandafter\def\csname PYG@tok@kn\endcsname{\let\PYG@bf=\textbf\def\PYG@tc##1{\textcolor[rgb]{0.00,0.50,0.00}{##1}}}
\expandafter\def\csname PYG@tok@cpf\endcsname{\let\PYG@it=\textit\def\PYG@tc##1{\textcolor[rgb]{0.25,0.50,0.50}{##1}}}
\expandafter\def\csname PYG@tok@kr\endcsname{\let\PYG@bf=\textbf\def\PYG@tc##1{\textcolor[rgb]{0.00,0.50,0.00}{##1}}}
\expandafter\def\csname PYG@tok@s\endcsname{\def\PYG@tc##1{\textcolor[rgb]{0.73,0.13,0.13}{##1}}}
\expandafter\def\csname PYG@tok@kp\endcsname{\def\PYG@tc##1{\textcolor[rgb]{0.00,0.50,0.00}{##1}}}
\expandafter\def\csname PYG@tok@w\endcsname{\def\PYG@tc##1{\textcolor[rgb]{0.73,0.73,0.73}{##1}}}
\expandafter\def\csname PYG@tok@kt\endcsname{\def\PYG@tc##1{\textcolor[rgb]{0.69,0.00,0.25}{##1}}}
\expandafter\def\csname PYG@tok@sc\endcsname{\def\PYG@tc##1{\textcolor[rgb]{0.73,0.13,0.13}{##1}}}
\expandafter\def\csname PYG@tok@sb\endcsname{\def\PYG@tc##1{\textcolor[rgb]{0.73,0.13,0.13}{##1}}}
\expandafter\def\csname PYG@tok@k\endcsname{\let\PYG@bf=\textbf\def\PYG@tc##1{\textcolor[rgb]{0.00,0.50,0.00}{##1}}}
\expandafter\def\csname PYG@tok@se\endcsname{\let\PYG@bf=\textbf\def\PYG@tc##1{\textcolor[rgb]{0.73,0.40,0.13}{##1}}}
\expandafter\def\csname PYG@tok@sd\endcsname{\let\PYG@it=\textit\def\PYG@tc##1{\textcolor[rgb]{0.73,0.13,0.13}{##1}}}

\def\PYGZbs{\char`\\}
\def\PYGZus{\char`\_}

\def\PYGZsh{\char`\#}

\def\PYGZhy{\char`\-}
\def\PYGZsq{\char`\'}
\def\PYGZdq{\char`\"}

\makeatother

\DeclareMathOperator*{\argmin}{arg\!\min}
\DeclareMathOperator*{\argmax}{arg\!\max}

\newcommand{\setmode}[1]{\def\mode{#1}}
\setmode{draft} 
\long\def\IGNORE#1{} \long\def\COMMENT#1{}
\def\authornote#1#2#3{{\textcolor{#2}{\textsl{\small#1:[*#3*]}}}}

\ifthenelse{\equal{\mode}{draft}} { 
	\newcommand{\zlnote}[1]{\authornote{ZLV}{red}{#1}} 
	\newcommand{\jdnote}[1]{\authornote{JDong}{blue}{#1}} 
} {}

\ifthenelse{\equal{\mode}{final}} {
	\newcommand{\zlnote}[1]{}
	\newcommand{\jdnote}[1]{}
	\typeout{************* MODE: Final}
} {}

\title{\LARGE \bf
miniSAM: A Flexible Factor Graph Non-linear Least Squares Optimization Framework
}

\author{Jing Dong$^{1}$ and Zhaoyang Lv$^{2}$
\thanks{*This work was mostly finished when both authors were PhD students at College of Computing, Georgia Institute of Technology, Atlanta, USA. We would like to thank Prof. Frank Dellaert and Dr. Mustafa Mukadam giving suggestions on this work. This work received no financial support.}%
\thanks{$^{1}${\tt\small thu.dongjing@gmail.com}}%
\thanks{$^{2}${\tt\small zhaoyang.lv@gatech.edu}}%
}

\begin{document}

\maketitle
\thispagestyle{empty}
\pagestyle{empty}

\begin{abstract}

Many problems in computer vision and robotics can be phrased as 
non-linear least squares optimization problems represented by \emph{factor graphs}, 
for example, simultaneous localization and mapping (SLAM), structure from motion (SfM), 
motion planning, and control.
We have developed an open-source C++/Python framework \emph{miniSAM}, 
for solving such factor graph based least squares problems.
Compared to most existing frameworks for least squares solvers, miniSAM has 
(1) full Python/NumPy API, which enables more agile development and 
easy binding with existing Python projects, and
(2) a wide list of sparse linear solvers, including CUDA enabled sparse linear solvers.
Our benchmarking results shows miniSAM offers comparable performances on 
various types of problems, with more flexible and smoother development experience.
\end{abstract}

\section{Introduction}

Solving non-linear least squares is important to many areas in robotics, including SLAM~\cite{Dellaert06ijrr}, SfM~\cite{Triggs00}, motion planning~\cite{Dong16rss}, and control~\cite{ta2014factor,mukadam2017approximately}.
Furthermore, researchers in these areas often use \emph{factor graphs}, a probabilistic graphical representation to model the non-linear least squares problem.
Dellaert and Kaess \cite{Dellaert06ijrr} first connected factor graphs to non-linear least squares, and the graph inference algorithms to sparse linear algebra algorithms.

There are existing libraries for solving non-linear least squares problems.
Existing widely used frameworks by SLAM and SfM communities include Ceres~\cite{ceressolver}, g2o~\cite{Kuemmerle11icra}, and GTSAM~\cite{Dellaert12tr}.
In particular, GTSAM uses factor graph to model the non-linear least square problems, and solves the problems using graphical algorithms rather than sparse linear algebra algorithms.
However, for performance reasons all existing frameworks are implemented in C++ and therefore have the disadvantage that they require complex C++ programing, especially when users merely want to define or customize loss functions.

We introduce a flexible, general and lightweight factor graph optimization framework \emph{miniSAM}\footnote[2]{https://github.com/dongjing3309/minisam}.
Like GTSAM, miniSAM uses factor graphs to model non-linear least square problems. 
The APIs and implementation of miniSAM are heavily inspired and influenced by GTSAM, but miniSAM is a much more lightweight framework, and that extends the flexibility of GTSAM as follows:
\begin{itemize}
\item Full Python/NumPy API, with the ability to define custom cost functions and optimizable manifolds to enable faster and easier prototyping.
\item A wide list of sparse linear solver choices, including CUDA supported GPU sparse linear solvers.
\item It is lightweight and requires minimal external dependencies, thus making it great for cross-platform compatibility. 
\end{itemize}

In this paper, we first give an introduction to non-linear least squares and the connection between sparse least squares and factor graphs. Then, we introduce the features and basic usage of miniSAM, finally we show benchmarking results of miniSAM on various SLAM problems.

\begin{figure}
\centering
\includegraphics[width=0.95\linewidth]{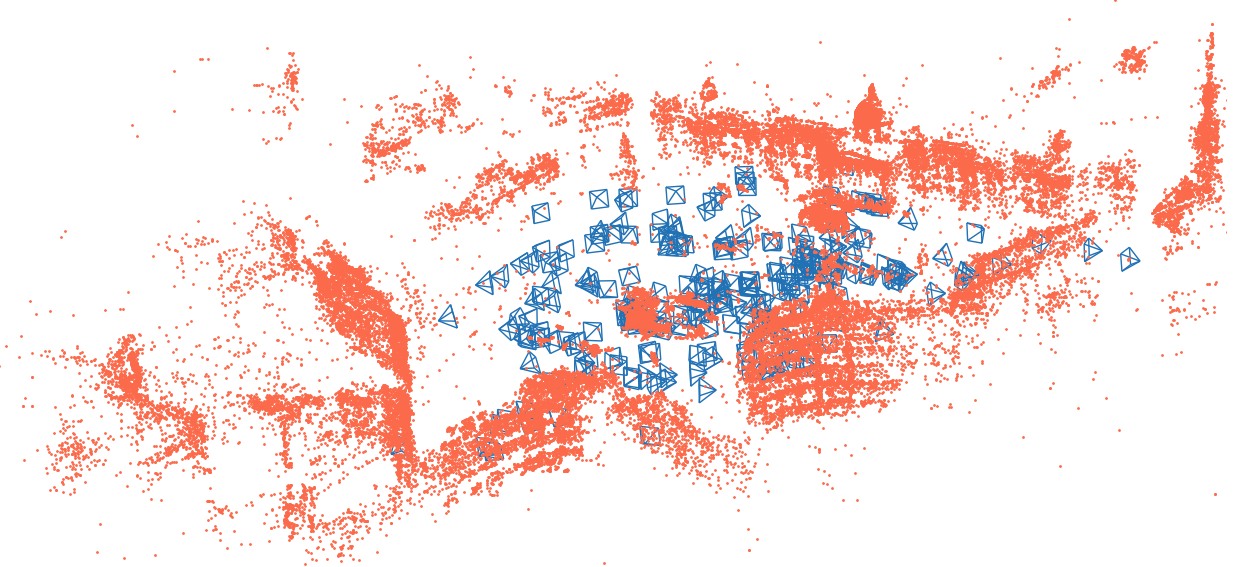}
\includegraphics[width=0.41\linewidth]{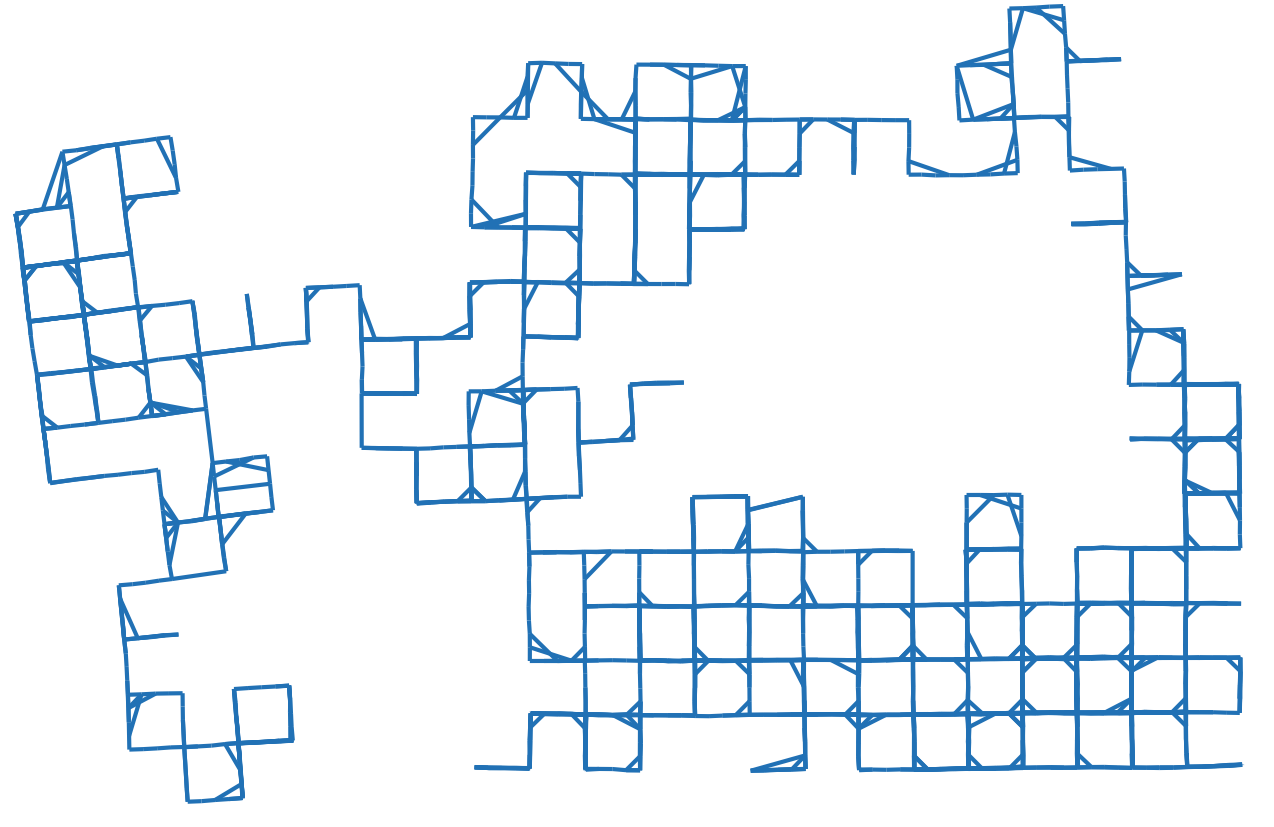}
\quad
\includegraphics[width=0.24\linewidth]{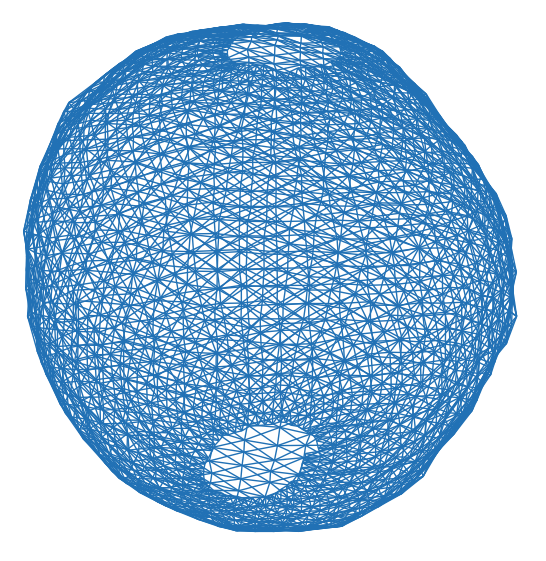}
\quad
\includegraphics[width=0.15\linewidth]{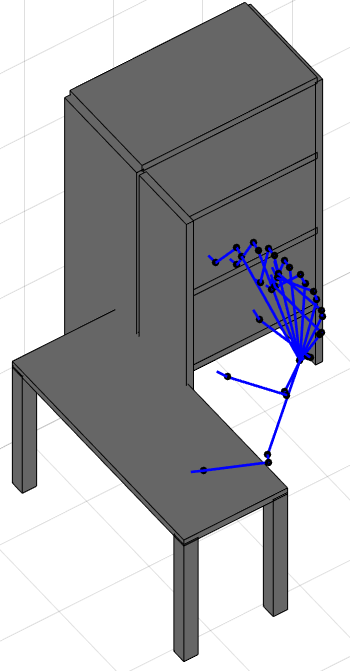}
\protect\caption{
Example problems solved by miniSAM. 
Top: bundle adjustment problem {\small\tt Trafalgar}~\cite{Agarwal10eccv}, camera poses are shown in blue and landmarks are shown in red.
Bottom (from left to right): 2D pose graph problem {\small\tt M3500}~\cite{carlone2015initialization}, 3D pose graph problem {\small\tt Sphere}~\cite{carlone2015initialization}, Barrett WAM arm motion planning problem~\cite{Dong16rss}.
\label{fig:examples}}
\end{figure}

\section{Introduction to Non-linear Least Squares and Factor Graphs}

\subsection{Non-linear Least Square Optimization}\label{sec:ls}

Non-linear least squares optimization is defined by 
\begin{equation}
x^{*} = \argmin_{x} \sum_i \rho_i \big( \parallel f_i(x) \parallel^{2}_{{\Sigma}_i} \big), \label{eq:ls}
\end{equation}
where $x \in \mathcal{M}$ is a point on a general $n$-dimensional manifold, $x^{*} \in \mathcal{M}$ is the solution, $f_i \in \mathbb{R}^m$ is a $m$-dimensional vector-valued error function, $\rho_i$ is a robust kernel function, and ${\Sigma}_i \in \mathbb{R}^{m \times m}$ is a covariance matrix. The Mahalanobis distance is defined by $\parallel v \parallel^{2}_{{\Sigma}} \doteq v^T {\Sigma}^{-1} v$ where $v \in \mathbb{R}^m$ and ${\Sigma}^{-1}$ is the information matrix. If we factorize the information matrix by Cholesky factorization ${\Sigma}^{-1} = {R}^T {R}$, where ${R}$ is upper triangular, we have
\begin{equation}
\parallel v \parallel^{2}_{{\Sigma}} = v^T {\Sigma}^{-1} v = v^T {R}^T {R} v
= \parallel {R} v  \parallel^{2}. \label{eq:mahalanobis}
\end{equation}
If we consider the simplified case where $\rho_i$ is identity and define $h_i(x) \doteq {R}_i f_i(x)$, then Eq.~(\ref{eq:ls}) is equivalent to
\begin{equation}
x^{*} = \argmin_{x} \sum_i \parallel h_i(x) \parallel^{2} \label{eq:ls_gaussian}
\end{equation}
as per Eq.~(\ref{eq:mahalanobis}). If we define a \emph{linearization point} $x_0 \in \mathcal{M}$, and the Jacobian matrix of $h_i(x)$
\begin{equation}
{J}_i \doteq \frac{\partial h_i(x)}{\partial x} \Big|_{x=x_0} \label{eq:jacobian}
\end{equation}
then the Taylor expansion is given by
\begin{equation}
h_i(x_0 + \Delta x) = h_i(x_0) + {J}_i \Delta x + O(\Delta x^2),  \label{eq:linearization_taylor}
\end{equation}
which we can use to solve the least square problem by searching a \emph{local} region near $x_0$, and find the solution by iteratively solving a \emph{linearized} least squares problem
\begin{equation}
\Delta x^{*} = \argmin_{\Delta x} \sum_i \parallel {J}_i \Delta x + h_i(x_0)  \parallel^{2}, \label{eq:linear_ls}
\end{equation}
where $\Delta x^{*} \in \mathbb{R}^n$, and the solution is updated by 
\begin{equation}
x^{*} = x_0 + \Delta x^{*}. \label{eq:linear_ls_solution}
\end{equation}
If $\mathcal{M}$ is simply a vector space $\mathbb{R}^n$ then
the above procedure is performed iteratively in general by setting $x_0$ of next iteration from $x^*$ of current iteration, until $x^*$ converges. 
Trust-region policies like Levenberg-Marquardt can be also applied when looking for $\Delta x^{*}$.

When $\mathcal{M}$ is a general manifold, we need to define a local coordinate chart of $\mathcal{M}$ near $x_0$, which is an invertible map between a local region of $\mathcal{M}$ around $x_0$ and the local Euclidean space, and also an operator $\oplus$ that maps a point in local Euclidean space back to $\mathcal{M}$. 
Thus Eq.~(\ref{eq:linear_ls_solution}) on general manifolds is
\begin{equation}
x^{*} = x_0 \oplus \Delta x^{*}. \label{eq:linear_ls_solution_manifold}
\end{equation}
A simple example of $\oplus$ is for the Euclidean space where it is simply the plus operator.

To solve the linear least squares problem in Eq.~(\ref{eq:linear_ls}), we first rewrite Eq.~(\ref{eq:linear_ls}) as
\begin{equation}
\Delta x^{*} = \argmin_{\Delta x} \parallel {J} \Delta x + b  \parallel^{2}, \label{eq:linear_ls_comb}
\end{equation}
where $J$ is defined by stacking all $J_i$ vertically, similarly $b$ is defined by stacking all $h_i(x_0)$ vertically.
Cholesky factorization is commonly used solve Eq.~(\ref{eq:linear_ls_comb}).  Since the solution of linear least squares problem in Eq.~(\ref{eq:linear_ls_comb}) is given by the normal equation
\begin{equation}
J^T J \Delta x^{*} = J^T b, \label{eq:normal_eq}
\end{equation}
we apply Cholesky factorization to symmetric $J^T J$, and we have $J^T J = R^T R$ where $R$ is upper triangular. 
Then solving Eq.~(\ref{eq:normal_eq}) is equivalent to solving both
\begin{align}
&R^T y = J^T b \label{eq:cholesky_1} \\
&R \Delta x^{*} = y \label{eq:cholesky_2}
\end{align}
in two steps, which can be both solved by back-substitution given that $R$ is triangular.
Other than Cholesky factorization, QR and SVD factorizations can be also used to solve Eq.~(\ref{eq:linear_ls_comb}), although with significantly slower speeds.
Iterative methods like pre-conditioned conjugate gradient (PCG) are also widely used to solve Eq.~(\ref{eq:normal_eq}), especially when $J^T J$ is very large.

\subsection{Connection between Factor Graphs and Sparse Least Squares}

Dellaert and Kaess~\cite{Dellaert06ijrr} have shown factor graphs have a tight connections with non-linear least square problems. 
A factor graph is a probabilistic graphical model, which represents a joint probability distribution of all factors
\begin{equation}
p(x) \propto \prod_i p_i(x_i), \label{eq:factor_graph}
\end{equation}
where $x_i \subseteq x$ is a subset of variables involved in factor $i$,
$p(x)$ is the overall distribution of the factor graph, and $p_i(x_i)$ is the distribution of each factor. 
The maximum a posteriori (MAP) estimate of the graph is
\begin{equation}
x^{*} = \argmax_{x} p(x) = \argmax_{x} \prod_i p_i(x_i). \label{eq:factor_graph_map}
\end{equation}
If we consider the case where each factor has Gaussian distribution on $f_i(x_i)$ with covariance $\Sigma_i$,
\begin{equation}
p_i(x_i) \propto \mathrm{exp} \big( - \frac{1}{2} \parallel f_i(x_i) \parallel^{2}_{{\Sigma}_i} \big), \label{eq:gaussian_factor}
\end{equation}
then MAP inference is
\begin{align}
x^{*} & = \argmax_{x} \prod_i p_i(x_i) = \argmax_{x} \mathrm{log} \big( \prod_i p_i(x_i) \big),  \\
& = \argmin_{x} \prod_i -\mathrm{log} \big( p_i(x_i) \big) = \argmin_{x} \sum_i \parallel f_i(x_i) \parallel^{2}_{{\Sigma}_i}. \label{eq:factor_graph_map_log}
\end{align}
The MAP inference problem in Eq.~(\ref{eq:factor_graph_map_log}) is converted to the same non-linear least squares optimization problem in Eq.~\ref{eq:ls}, which can be solved following the same steps in Section~\ref{sec:ls}.

There are several advantages of using factor graph to model the non-linear least squares problem in SLAM.
Factor graphs encode the probabilistic nature of the problem, and easily visualize the underlying sparsity of most SLAM problems since for most (if not all) factors $x_i$ are very small sets.
We give an example in the next section, which clearly visualizes this sparsity in a factor graph.

\subsection{Example: A Pose Graph}\label{sec:pg5_example}

\begin{figure}
\centering
\begin{subfigure}[b]{0.25\textwidth}
\centering
\includegraphics[width=1\linewidth]{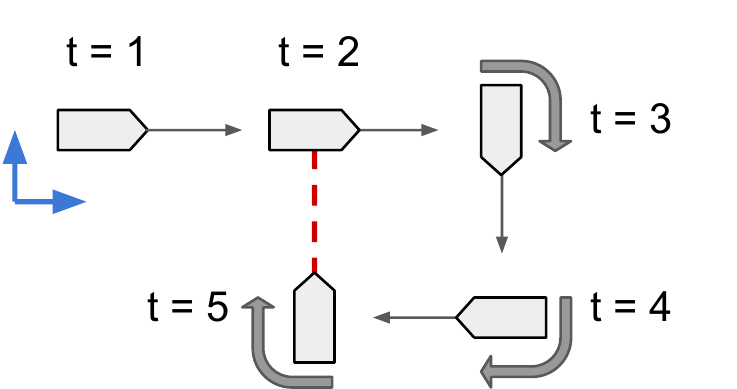}
\caption{Problem}\label{fig:pose_graph_problem}
\end{subfigure}
\begin{subfigure}[b]{0.4\textwidth}
\centering
\includegraphics[width=1\linewidth]{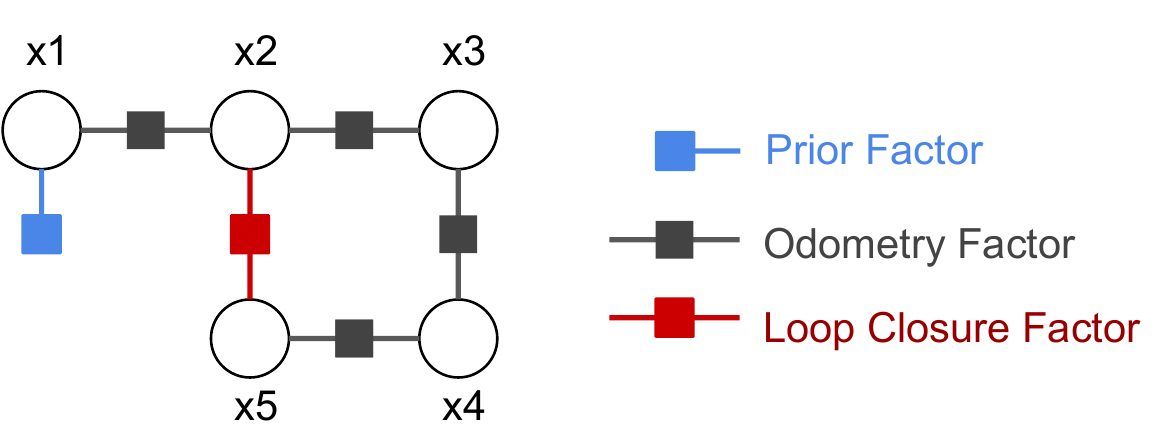}
\caption{Factor graph}\label{fig:pose_graph_fg}
\end{subfigure}
\protect\caption{
An example 2D pose graph problem, with pose variables (shown as white circles), a prior (shown as a blue factor), four odometry measurements (shown as black factors) and a single loop closure measurement (shown as a red factor).
\label{fig:pose_graph_example}}
\end{figure}

Here we give a simple example of using factor graph to solve a small pose graph problem.
The problem is shown in Fig.~\ref{fig:pose_graph_problem}, where a vehicle moves forward on a 2D plane, makes a 270 degrees right turn, and has a relative pose loop closure measurement which is shown in red. 
If we want to estimate the vehicle's poses at times $t=1,2,3,4,5$, we define the system's state variables
\begin{equation}
x = \{x_1, x_2, x_3, x_4, x_5\}, \label{eq:state_pg5}
\end{equation}
where $x_i \in SE(2), i=1,2,3,4,5$ is the vehicle pose at $t=i$.
Then the factor graph models the pose graph problem as
\begin{equation}
p(x) \propto \underbrace{p(x_1)}_\text{prior}\underbrace{p(x_1, x_2)p(x_2, x_3)p(x_3, x_4)p(x_4, x_5)}_\text{odometry}\underbrace{p(x_2, x_5)}_\text{loop closure} \label{eq:factor_graph_pg5}
\end{equation}
which is shown in Fig.~\ref{fig:pose_graph_fg}.
As shown in Eq.~(\ref{eq:factor_graph_pg5}) and Fig.~\ref{fig:pose_graph_fg}, there are three types of factors: (1) A prior factor, which gives a prior distribution on first pose, and locks the solution to a world coordinate frame. (2) Odometry factors, which encode the relative poses odometry measurements between $t=i$ and $t=i+1$. (3) A loop closure factor, which encodes the relative poses measurement between $t=2$ and $t=5$.

The sparsity of the example problem is clearly shown by the factor graph: all the factors are unary or binary. This is actually true for all pose graph optimization problems. We can further show the sparsity of the underlying linear system we solve in Eq.~(\ref{eq:linear_ls_comb}) and Eq.~(\ref{eq:normal_eq})
\begin{align}
J =& \begin{pmatrix} 
J_{11} & & & & \\  \hdashline
J_{21} & J_{22} & & & \\
 & J_{32} & J_{33} & & \\
 & & J_{43} & J_{44} & \\
 & & & J_{54} & J_{55} \\  \hdashline
 & J_{62} & & & J_{65} \\
\end{pmatrix}
\begin{matrix*}[l]
\text{\footnotesize prior}\\
\\
\text{\footnotesize odometry}\\
\\
\\
\text{\footnotesize loop closure}\\
\end{matrix*},
\\
H = J^T J =& \begin{pmatrix} 
H_{11} & H_{12} & & & \\
H_{21} & H_{22} & H_{23} & & H_{25}\\
 & H_{32} & H_{33} & H_{34} & \\
 & & H_{43} & H_{44} & H_{45}\\
 & H_{52} & & H_{54} & H_{55} \\
\end{pmatrix}.
\end{align}
We can clearly see both $J$ and $J^T J$ have block-wise sparse structures.

\section{Solving Non-linear Least Square Problems with miniSAM}

In this section we give some details about our miniSAM library and the basic introduction of how to use miniSAM.
miniSAM is implemented with C++11, and provides both native C++ API and Python API provided by pybind11~\cite{pybind11}.
miniSAM is a very lightweight library: the core implementation has only 8k lines of C++ code, plus 4k lines of test code and 2k lines of Python wrapper code. Also miniSAM requires minimal external dependencies (only Eigen~\cite{eigenweb} linear algebra library is required).
This makes miniSAM great for cross-platform compatibility. 
Currently miniSAM compiles with most major compliers (GCC, Clang, Microsoft Visual C++) on most major OSs (Linux, macOS, Windows).

One of the most important features of miniSAM is its high flexibility with the Python interface, which provides the ability to create custom factors and manifolds directly in Python. Although most existing frameworks provide (or by third-parties) binding to script languages (like Python and MATLAB) that enable them to define graph structures, they all lack ability to directly define factors and manifolds in script languages. To use custom factors or manifolds in script languages, users need to first define them in C++, then bind the interfaces to script languages, which is inconvenient.
Table~\ref{table:features} gives a comparison of multiple features in C++/Python/MATLAB between miniSAM and existing frameworks.

In the rest of this section we discuss three major use cases of miniSAM: how to define and solve factor graphs/least square optimization problems, create custom factors/cost functions, and create custom optimizable manifolds.

\begin{table}
\caption{Features comparison across frameworks.\label{table:features}}
\footnotesize
\begin{tabular}{lcccc}
\hline
& Ceres & g2o & GTSAM & miniSAM \\
\hline
C++: Custom cost/factor & \textbullet & \textbullet & \textbullet & \textbullet \\
C++: Numerical jacobian & \textbullet & \textbullet & & \textbullet \\
C++: Auto-diff jacobian & \textbullet & \textbullet & \textbullet & \\
C++: Custom manifold & \textbullet & \textbullet & \textbullet & \textbullet \\
MATLAB: API binding & & & \textbullet & \\
Python: API binding & \textbullet$^1$ & \textbullet$^1$ & \textbullet & \textbullet \\
Python: Custom cost/factor & & & & \textbullet \\
Python: Numerical jacobian & & & & \textbullet \\
Python: Auto-diff jacobian & & & & \\
Python: Custom manifold & & & & \textbullet \\
\hline
\end{tabular}
$^1$by third-party.\\
\end{table}

\subsection{The Pose Graph Example}

Here we give an example on how to use miniSAM to solve the pose graph example discussed in Section~\ref{sec:pg5_example}. 
Example Python code solving this pose graph example in Fig.~\ref{fig:pose_graph_example} is in Snippet~1 in the Appendix.

In the first step we construct the factor graph. In miniSAM data structure {\small\tt FactorGraph} is used as the container for factor graphs.
In miniSAM each variable is indexed by a \emph{key}, which is defined by a character and an unsigned integer (e.g. $x_1$).
Each factor has its key list that indicates the connected variables, and its loss function that has covariance $\Sigma_i$ and optional robust kernel $\rho_i$ (Cauchy and Huber robust loss functions have been implemented in miniSAM).
In the pose graph example two types of factors are used: unary {\small\tt PriorFactor} and binary {\small\tt BetweenFactor}.

In the second step we provide the initial variable values as the linearization point. 
In miniSAM variable values are stored in structure {\small\tt Variables}, where each variable is indexed by its key. 
Finally, we call a non-linear least square solver (like Levenberg-Marquardt) to solve the problem. Result variables are returned in a {\small\tt Variables} structure with status code.

\subsection{Define Factors}

Here we discuss how to define a new factor in miniSAM. 
As mentioned defining a new factor can be done in both C++ and Python in miniSAM, by inheriting from {\small\tt Factor} base class.
The implementation of a factor class includes an error function {\small\tt error()} that defines $f_i(x_i)$, which returns a {\small\tt Eigen::VectorXd} in C++, or a NumPy array in Python.
And Jacobian matrices function {\small\tt jacobians()} that defines ${\partial f_i(x_i)} / {\partial x_i}$ for each variable in $x_i$, which return a {\small\tt std::vector<Eigen::MatrixXd>} in C++, or a list of NumPy matrices in Python.
We show an example prior factor on $SE(2)$ in Python in Snippet~2.

Analytic Jacobians ${\partial f_i(x_i)} / {\partial x_i}$ is usually quite complex for non-trivial factors, and is the main bottleneck for faster prototyping. 
miniSAM provides a solution by inheriting from {\small\tt NumericalFactor} base class, numerical ${\partial f_i(x_i)} / {\partial x_i}$ through finite differencing will be evaluated during optimization, thus saving developer's time deriving analytic Jacobians.
We leave automatic differentiation for Jacobian evaluation as future work.

\subsection{Define Optimizable Manifolds}

miniSAM already has build-in support for optimizing various commonly used manifold types in C++ and Python, including Eigen vector types in C++, NumPy array in Python, and Lie groups $SO(2)$, $SE(2)$, $SO(3)$, $SE(3)$ and $Sim(3)$ (implementations provided by Sophus library~\cite{sophus}), which are commonly used in SLAM and robotics problems.

We can also customize manifold properties of any C++ or Python class for miniSAM.
In Python this is done by defining manifold-related member functions, including {\small\tt dim()} function returns manifold dimensionality, and {\small\tt local()} and {\small\tt retract()} functions defines the local coordinate chart.
An example of defining a vector space manifold $\mathbb{R}^2$ in Python is in Snippet~3.
In C++ we use a non-intrusive technique called \emph{traits}, which is a specialization of template {\small\tt minisam::traits<>} for the type we are adding manifold properties. 
Using traits to define manifold properties has two advantages: (1) optimizing a class without modifying it, or even without knowing details of implementation (e.g. adding miniSAM optimization support for third-party C/C++ types), (2) making optimizing primitive type (like float/double) possible.

%

\section{Experiments}

To test the performance of miniSAM, we run a benchmark on multiple problems of different types and scales, and compare with multiple existing frameworks.
We choose three SLAM and SfM problem for benchmarking, from small to large. 
\begin{itemize}
\item 2D pose graph problem {\small\tt M3500}~\cite{carlone2015initialization}, which contains 3500 2D poses and 5453 energy edges.
\item 3D pose graph problem {\small\tt Torus}~\cite{carlone2015initialization}, which contains 5000 3D poses and 9048 energy edges.
\item Bundle adjustment problem {\small\tt Dubrovnik}~\cite{Agarwal10eccv}, which contains 356 camera poses, 226730 landmarks and 1255268 image measurements. 
\end{itemize}
For all problems we use Levenberg-Marquardt algorithm to solve, and fix the number of iterations to 5.

We run the benchmark with the following frameworks and sparse linear solvers
\begin{itemize}
\item Ceres~\cite{ceressolver} with Eigen simplicial LDLT solver, and CHOLMOD~\cite{chen2008algorithm} Cholesky solver.
\item g2o~\cite{Kuemmerle11icra} with Eigen simplicial LDLT solver, and CHOLMOD Cholesky solver.
\item GTSAM~\cite{Dellaert12tr} with built-in multi-frontal and sequential graph elimination solvers.
\item miniSAM with Eigen simplicial LDLT solver, CHOLMOD Cholesky solver, and CUDA cuSOLVER GPGPU Cholesky solver.
\end{itemize}
%


\begin{table}
\caption{Optimization times in second of different frameworks with different sparse linear solvers, grouped by single-thread or multi-thread.\label{table:timing_results}}
\small
\begin{tabular}{lcccc}
\hline
& 2D-PG & 3D-PG & BA \\[1pt]
\hline
Ceres + Eigen LDLT & 0.090 & 2.735 & 54.96 \\[1pt]
g2o + Eigen LDLT & 0.059 & 2.697 & 63.66 \\[1pt]
GTSAM + Multifrontal Cholesky & 0.228 & 2.002 & 83.67 \\[1pt]
GTSAM + Sequential Cholesky & 0.207 & 2.836 & 83.85 \\[1pt]
miniSAM + Eigen LDLT & 0.088 & 3.341 & 64.38 \\[1pt]
\hline
Ceres + CHOLMOD & 0.080 & 0.941 & 28.17 \\[1pt]
g2o + CHOLMOD & 0.064 &  0.821 & 35.68 \\[1pt]
miniSAM + CHOLMOD & 0.090 & 1.107 & 39.24 \\[1pt]
miniSAM + cuSOLVER Cholesky & 0.458 & 1.791 & 49.77 \\[1pt]
\hline
\end{tabular}
\end{table}

For miniSAM, all factors and manifolds are implemented natively in C++.
All frameworks in benchmarking are compiled in single-thread, except CHOLMOD and CUDA cuSOLVER solvers are compiled in multi-thread (using all 12 available CPU threads during benchmarking, and GPU is used with CUDA).
The benchmarking is performed on a computer with Intel Core i7-6850K CPU, 128 GB memory, and a NVIDIA TITAN X GPU with 12GB graphic memory. 
The results are shown in Table.~\ref{table:timing_results}, and are grouped by single-thread or multi-thread.

We can see in Table.~\ref{table:timing_results} that when the same sparse linear solver is used, miniSAM has slightly worse runtime compare to Ceres and g2o, but (except for 3D pose graph case) has better runtime compared to GTSAM, which does not use third-party sparse linear solvers.
The extra overhead of miniSAM compare to Ceres and g2o are mainly due to two major miniSAM design choices:
\begin{itemize}
\item miniSAM avoids using any compile-time array or matrix, and all internal vectors and matrices are dynamically allocated. The use of dynamic size arrays involves extra memory allocation overhead and forbids any compile-time optimization by modern CPU SIMD instructions. 
\item miniSAM avoids using any raw pointer and manual memory management.
\end{itemize}
The reason to make above design choices is that to make miniSAM have a Python API consistent with C++ API, and to make Python interface possible to implement, since Python does not have machinery to support template programming or explicit memory management.

We also found CUDA cuSOLVER is not as fast as CHOLMOD CPU solver when using all 12 available CPU threads, and it is particularly slow on small problems. 
Finally, CUDA cuSOLVER has an one-time launch delay of about 350ms, once per executable launch. Given such circumstances using CUDA cuSOLVER is currently only good for large problems. 


\section{Conclusion}

We gave a brief introduction to miniSAM, our non-linear least squares optimization library.
We demonstrate the basic usage of miniSAM, show its flexibility in fast prototyping in Python, and its performance in benchmarking of multiple types of problems in SLAM and robotics applications.
We recognize miniSAM has a relatively small performance loss compared to other state-of-the-art frameworks, mostly due to miniSAM's design to adapt Python API, so currently miniSAM is not great for performance-critical applications. But hopefully we can solve the problem in the future by porting better sparse linear solvers (like GPU-enabled iterative solver) to mitigate this issue.

\bibliographystyle{ieeetr}
\bibliography{refs}


\begin{figure*}
\section*{Appendix: Python Example Code Snippets}
\vspace*{8mm}
\begin{tcolorbox}[title=Snippet 1. A pose graph optimization in Python,boxrule=0.5pt,arc=.2em,boxsep=1pt,left=2pt,right=2pt,top=2pt,bottom=2pt]

\begin{scriptsize}
\begin{Verbatim}[commandchars=\\\{\}]
\PYG{k+kn}{import} \PYG{n+nn}{numpy} \PYG{k+kn}{as} \PYG{n+nn}{np}
\PYG{k+kn}{from} \PYG{n+nn}{minisam} \PYG{k+kn}{import} \PYG{o}{*}
\PYG{k+kn}{from} \PYG{n+nn}{minisam.sophus} \PYG{k+kn}{import} \PYG{o}{*}

\PYG{c+c1}{\PYGZsh{} build factor graph for least square problem}
\PYG{n}{graph} \PYG{o}{=} \PYG{n}{FactorGraph}\PYG{p}{()}
\PYG{n}{loss} \PYG{o}{=} \PYG{n}{DiagonalLoss}\PYG{o}{.}\PYG{n}{Sigmas}\PYG{p}{(}\PYG{n}{np}\PYG{o}{.}\PYG{n}{array}\PYG{p}{([}\PYG{l+m+mf}{1.0}\PYG{p}{,} \PYG{l+m+mf}{1.0}\PYG{p}{,} \PYG{l+m+mf}{0.1}\PYG{p}{]))} \PYG{c+c1}{\PYGZsh{} loss function of sensor measurement model}
\PYG{n}{graph}\PYG{o}{.}\PYG{n}{add}\PYG{p}{(}\PYG{n}{PriorFactor}\PYG{p}{(}\PYG{n}{key}\PYG{p}{(}\PYG{l+s+s1}{\PYGZsq{}x\PYGZsq{}}\PYG{p}{,} \PYG{l+m+mi}{1}\PYG{p}{),} \PYG{n}{SE2}\PYG{p}{(}\PYG{n}{SO2}\PYG{p}{(}\PYG{l+m+mi}{0}\PYG{p}{),} \PYG{n}{np}\PYG{o}{.}\PYG{n}{array}\PYG{p}{([}\PYG{l+m+mi}{0}\PYG{p}{,} \PYG{l+m+mi}{0}\PYG{p}{])),} \PYG{n}{loss}\PYG{p}{))} \PYG{c+c1}{\PYGZsh{} prior as first pose}
\PYG{n}{graph}\PYG{o}{.}\PYG{n}{add}\PYG{p}{(}\PYG{n}{BetweenFactor}\PYG{p}{(}\PYG{n}{key}\PYG{p}{(}\PYG{l+s+s1}{\PYGZsq{}x\PYGZsq{}}\PYG{p}{,} \PYG{l+m+mi}{1}\PYG{p}{),} \PYG{n}{key}\PYG{p}{(}\PYG{l+s+s1}{\PYGZsq{}x\PYGZsq{}}\PYG{p}{,} \PYG{l+m+mi}{2}\PYG{p}{),} \PYG{n}{SE2}\PYG{p}{(}\PYG{n}{SO2}\PYG{p}{(}\PYG{l+m+mi}{0}\PYG{p}{),} \PYG{n}{np}\PYG{o}{.}\PYG{n}{array}\PYG{p}{([}\PYG{l+m+mi}{5}\PYG{p}{,} \PYG{l+m+mi}{0}\PYG{p}{])),} \PYG{n}{loss}\PYG{p}{))} \PYG{c+c1}{\PYGZsh{} odometry measurements}
\PYG{n}{graph}\PYG{o}{.}\PYG{n}{add}\PYG{p}{(}\PYG{n}{BetweenFactor}\PYG{p}{(}\PYG{n}{key}\PYG{p}{(}\PYG{l+s+s1}{\PYGZsq{}x\PYGZsq{}}\PYG{p}{,} \PYG{l+m+mi}{2}\PYG{p}{),} \PYG{n}{key}\PYG{p}{(}\PYG{l+s+s1}{\PYGZsq{}x\PYGZsq{}}\PYG{p}{,} \PYG{l+m+mi}{3}\PYG{p}{),} \PYG{n}{SE2}\PYG{p}{(}\PYG{n}{SO2}\PYG{p}{(}\PYG{o}{\PYGZhy{}}\PYG{l+m+mf}{3.14}\PYG{o}{/}\PYG{l+m+mi}{2}\PYG{p}{),} \PYG{n}{np}\PYG{o}{.}\PYG{n}{array}\PYG{p}{([}\PYG{l+m+mi}{5}\PYG{p}{,} \PYG{l+m+mi}{0}\PYG{p}{])),} \PYG{n}{loss}\PYG{p}{))}
\PYG{n}{graph}\PYG{o}{.}\PYG{n}{add}\PYG{p}{(}\PYG{n}{BetweenFactor}\PYG{p}{(}\PYG{n}{key}\PYG{p}{(}\PYG{l+s+s1}{\PYGZsq{}x\PYGZsq{}}\PYG{p}{,} \PYG{l+m+mi}{3}\PYG{p}{),} \PYG{n}{key}\PYG{p}{(}\PYG{l+s+s1}{\PYGZsq{}x\PYGZsq{}}\PYG{p}{,} \PYG{l+m+mi}{4}\PYG{p}{),} \PYG{n}{SE2}\PYG{p}{(}\PYG{n}{SO2}\PYG{p}{(}\PYG{o}{\PYGZhy{}}\PYG{l+m+mf}{3.14}\PYG{o}{/}\PYG{l+m+mi}{2}\PYG{p}{),} \PYG{n}{np}\PYG{o}{.}\PYG{n}{array}\PYG{p}{([}\PYG{l+m+mi}{5}\PYG{p}{,} \PYG{l+m+mi}{0}\PYG{p}{])),} \PYG{n}{loss}\PYG{p}{))}
\PYG{n}{graph}\PYG{o}{.}\PYG{n}{add}\PYG{p}{(}\PYG{n}{BetweenFactor}\PYG{p}{(}\PYG{n}{key}\PYG{p}{(}\PYG{l+s+s1}{\PYGZsq{}x\PYGZsq{}}\PYG{p}{,} \PYG{l+m+mi}{4}\PYG{p}{),} \PYG{n}{key}\PYG{p}{(}\PYG{l+s+s1}{\PYGZsq{}x\PYGZsq{}}\PYG{p}{,} \PYG{l+m+mi}{5}\PYG{p}{),} \PYG{n}{SE2}\PYG{p}{(}\PYG{n}{SO2}\PYG{p}{(}\PYG{o}{\PYGZhy{}}\PYG{l+m+mf}{3.14}\PYG{o}{/}\PYG{l+m+mi}{2}\PYG{p}{),} \PYG{n}{np}\PYG{o}{.}\PYG{n}{array}\PYG{p}{([}\PYG{l+m+mi}{5}\PYG{p}{,} \PYG{l+m+mi}{0}\PYG{p}{])),} \PYG{n}{loss}\PYG{p}{))}
\PYG{n}{graph}\PYG{o}{.}\PYG{n}{add}\PYG{p}{(}\PYG{n}{BetweenFactor}\PYG{p}{(}\PYG{n}{key}\PYG{p}{(}\PYG{l+s+s1}{\PYGZsq{}x\PYGZsq{}}\PYG{p}{,} \PYG{l+m+mi}{5}\PYG{p}{),} \PYG{n}{key}\PYG{p}{(}\PYG{l+s+s1}{\PYGZsq{}x\PYGZsq{}}\PYG{p}{,} \PYG{l+m+mi}{2}\PYG{p}{),} \PYG{n}{SE2}\PYG{p}{(}\PYG{n}{SO2}\PYG{p}{(}\PYG{o}{\PYGZhy{}}\PYG{l+m+mf}{3.14}\PYG{o}{/}\PYG{l+m+mi}{2}\PYG{p}{),} \PYG{n}{np}\PYG{o}{.}\PYG{n}{array}\PYG{p}{([}\PYG{l+m+mi}{5}\PYG{p}{,} \PYG{l+m+mi}{0}\PYG{p}{])),} \PYG{n}{loss}\PYG{p}{))} \PYG{c+c1}{\PYGZsh{} loop closure}

\PYG{c+c1}{\PYGZsh{} variables initial guess, with random added\PYGZhy{}on noise}
\PYG{n}{init\PYGZus{}values} \PYG{o}{=} \PYG{n}{Variables}\PYG{p}{()}
\PYG{n}{init\PYGZus{}values}\PYG{o}{.}\PYG{n}{add}\PYG{p}{(}\PYG{n}{key}\PYG{p}{(}\PYG{l+s+s1}{\PYGZsq{}x\PYGZsq{}}\PYG{p}{,} \PYG{l+m+mi}{1}\PYG{p}{),} \PYG{n}{SE2}\PYG{p}{(}\PYG{n}{SO2}\PYG{p}{(}\PYG{l+m+mf}{0.2}\PYG{p}{),} \PYG{n}{np}\PYG{o}{.}\PYG{n}{array}\PYG{p}{([}\PYG{l+m+mf}{0.2}\PYG{p}{,} \PYG{o}{\PYGZhy{}}\PYG{l+m+mf}{0.3}\PYG{p}{])))}
\PYG{n}{init\PYGZus{}values}\PYG{o}{.}\PYG{n}{add}\PYG{p}{(}\PYG{n}{key}\PYG{p}{(}\PYG{l+s+s1}{\PYGZsq{}x\PYGZsq{}}\PYG{p}{,} \PYG{l+m+mi}{2}\PYG{p}{),} \PYG{n}{SE2}\PYG{p}{(}\PYG{n}{SO2}\PYG{p}{(}\PYG{o}{\PYGZhy{}}\PYG{l+m+mf}{0.1}\PYG{p}{),} \PYG{n}{np}\PYG{o}{.}\PYG{n}{array}\PYG{p}{([}\PYG{l+m+mf}{5.1}\PYG{p}{,} \PYG{l+m+mf}{0.3}\PYG{p}{])))}
\PYG{n}{init\PYGZus{}values}\PYG{o}{.}\PYG{n}{add}\PYG{p}{(}\PYG{n}{key}\PYG{p}{(}\PYG{l+s+s1}{\PYGZsq{}x\PYGZsq{}}\PYG{p}{,} \PYG{l+m+mi}{3}\PYG{p}{),} \PYG{n}{SE2}\PYG{p}{(}\PYG{n}{SO2}\PYG{p}{(}\PYG{o}{\PYGZhy{}}\PYG{l+m+mf}{3.14}\PYG{o}{/}\PYG{l+m+mi}{2} \PYG{o}{\PYGZhy{}} \PYG{l+m+mf}{0.2}\PYG{p}{),} \PYG{n}{np}\PYG{o}{.}\PYG{n}{array}\PYG{p}{([}\PYG{l+m+mf}{9.9}\PYG{p}{,} \PYG{o}{\PYGZhy{}}\PYG{l+m+mf}{0.1}\PYG{p}{])))}
\PYG{n}{init\PYGZus{}values}\PYG{o}{.}\PYG{n}{add}\PYG{p}{(}\PYG{n}{key}\PYG{p}{(}\PYG{l+s+s1}{\PYGZsq{}x\PYGZsq{}}\PYG{p}{,} \PYG{l+m+mi}{4}\PYG{p}{),} \PYG{n}{SE2}\PYG{p}{(}\PYG{n}{SO2}\PYG{p}{(}\PYG{o}{\PYGZhy{}}\PYG{l+m+mf}{3.14} \PYG{o}{+} \PYG{l+m+mf}{0.1}\PYG{p}{),} \PYG{n}{np}\PYG{o}{.}\PYG{n}{array}\PYG{p}{([}\PYG{l+m+mf}{10.2}\PYG{p}{,} \PYG{o}{\PYGZhy{}}\PYG{l+m+mf}{5.0}\PYG{p}{])))}
\PYG{n}{init\PYGZus{}values}\PYG{o}{.}\PYG{n}{add}\PYG{p}{(}\PYG{n}{key}\PYG{p}{(}\PYG{l+s+s1}{\PYGZsq{}x\PYGZsq{}}\PYG{p}{,} \PYG{l+m+mi}{5}\PYG{p}{),} \PYG{n}{SE2}\PYG{p}{(}\PYG{n}{SO2}\PYG{p}{(}\PYG{l+m+mf}{3.14}\PYG{o}{/}\PYG{l+m+mi}{2} \PYG{o}{\PYGZhy{}} \PYG{l+m+mf}{0.1}\PYG{p}{),} \PYG{n}{np}\PYG{o}{.}\PYG{n}{array}\PYG{p}{([}\PYG{l+m+mf}{5.1}\PYG{p}{,} \PYG{o}{\PYGZhy{}}\PYG{l+m+mf}{5.1}\PYG{p}{])))}

\PYG{c+c1}{\PYGZsh{} solve least square optimization by Levenberg\PYGZhy{}Marquardt algorithm}
\PYG{n}{opt} \PYG{o}{=} \PYG{n}{LevenbergMarquardtOptimizer}\PYG{p}{()}
\PYG{n}{result\PYGZus{}values} \PYG{o}{=} \PYG{n}{Variables}\PYG{p}{()} \PYG{c+c1}{\PYGZsh{} results}
\PYG{n}{status} \PYG{o}{=} \PYG{n}{opt}\PYG{o}{.}\PYG{n}{optimize}\PYG{p}{(}\PYG{n}{graph}\PYG{p}{,} \PYG{n}{init\PYGZus{}values}\PYG{p}{,} \PYG{n}{result\PYGZus{}values}\PYG{p}{)}
\PYG{k}{if} \PYG{n}{status} \PYG{o}{!=} \PYG{n}{NonlinearOptimizationStatus}\PYG{o}{.}\PYG{n}{SUCCESS}\PYG{p}{:}
    \PYG{k}{print}\PYG{p}{(}\PYG{l+s+s2}{\PYGZdq{}optimization error :\PYGZdq{}}\PYG{p}{,} \PYG{n}{status}\PYG{p}{)}
\end{Verbatim}
\end{scriptsize}

\end{tcolorbox}

\vspace*{5mm}

\begin{tcolorbox}[title=Snippet 2. A minimal Python prior factor example on SE(2),boxrule=0.5pt,arc=.2em,boxsep=1pt,left=2pt,right=2pt,top=2pt,bottom=2pt]

\begin{scriptsize}
\begin{Verbatim}[commandchars=\\\{\}]
\PYG{k+kn}{import} \PYG{n+nn}{numpy} \PYG{k+kn}{as} \PYG{n+nn}{np}
\PYG{k+kn}{from} \PYG{n+nn}{minisam} \PYG{k+kn}{import} \PYG{o}{*}

\PYG{c+c1}{\PYGZsh{} python implementation of prior factor on SE2}
\PYG{k}{class} \PYG{n+nc}{PyPriorFactorSE2}\PYG{p}{(}\PYG{n}{Factor}\PYG{p}{):} \PYG{c+c1}{\PYGZsh{} or inherit from NumericalFactor}
    \PYG{c+c1}{\PYGZsh{} constructor}
    \PYG{k}{def} \PYG{n+nf}{\PYGZus{}\PYGZus{}init\PYGZus{}\PYGZus{}}\PYG{p}{(}\PYG{n+nb+bp}{self}\PYG{p}{,} \PYG{n}{key}\PYG{p}{,} \PYG{n}{prior}\PYG{p}{,} \PYG{n}{loss}\PYG{p}{):}
        \PYG{n}{Factor}\PYG{o}{.}\PYG{n}{\PYGZus{}\PYGZus{}init\PYGZus{}\PYGZus{}}\PYG{p}{(}\PYG{n+nb+bp}{self}\PYG{p}{,} \PYG{l+m+mi}{3}\PYG{p}{,} \PYG{p}{[}\PYG{n}{key}\PYG{p}{],} \PYG{n}{loss}\PYG{p}{)}
        \PYG{n+nb+bp}{self}\PYG{o}{.}\PYG{n}{prior\PYGZus{}} \PYG{o}{=} \PYG{n}{prior}
    \PYG{c+c1}{\PYGZsh{} make a deep copy}
    \PYG{k}{def} \PYG{n+nf}{copy}\PYG{p}{(}\PYG{n+nb+bp}{self}\PYG{p}{):}
        \PYG{k}{return} \PYG{n}{PyPriorFactorSE2}\PYG{p}{(}\PYG{n+nb+bp}{self}\PYG{o}{.}\PYG{n}{keys}\PYG{p}{()[}\PYG{l+m+mi}{0}\PYG{p}{],} \PYG{n+nb+bp}{self}\PYG{o}{.}\PYG{n}{prior\PYGZus{}}\PYG{p}{,} \PYG{n+nb+bp}{self}\PYG{o}{.}\PYG{n}{lossFunction}\PYG{p}{())}
    \PYG{c+c1}{\PYGZsh{} error vector}
    \PYG{k}{def} \PYG{n+nf}{error}\PYG{p}{(}\PYG{n+nb+bp}{self}\PYG{p}{,} \PYG{n}{variables}\PYG{p}{):}
        \PYG{n}{curr\PYGZus{}pose} \PYG{o}{=} \PYG{n}{variables}\PYG{o}{.}\PYG{n}{at}\PYG{p}{(}\PYG{n+nb+bp}{self}\PYG{o}{.}\PYG{n}{keys}\PYG{p}{()[}\PYG{l+m+mi}{0}\PYG{p}{])} \PYG{c+c1}{\PYGZsh{} current variable}
        \PYG{k}{return} \PYG{p}{(}\PYG{n+nb+bp}{self}\PYG{o}{.}\PYG{n}{prior\PYGZus{}}\PYG{o}{.}\PYG{n}{inverse}\PYG{p}{()} \PYG{o}{*} \PYG{n}{curr\PYGZus{}pose}\PYG{p}{)}\PYG{o}{.}\PYG{n}{log}\PYG{p}{()}
    \PYG{c+c1}{\PYGZsh{} jacobians, not needed if inherit from NumericalFactor}
    \PYG{k}{def} \PYG{n+nf}{jacobians}\PYG{p}{(}\PYG{n+nb+bp}{self}\PYG{p}{,} \PYG{n}{variables}\PYG{p}{):}
        \PYG{k}{return} \PYG{p}{[}\PYG{n}{np}\PYG{o}{.}\PYG{n}{eye}\PYG{p}{(}\PYG{l+m+mi}{3}\PYG{p}{)]}
\end{Verbatim}
\end{scriptsize}

\end{tcolorbox}

\vspace*{5mm}

\begin{tcolorbox}[title=Snippet 3. A minimal Python 2D point optimizable manifold,boxrule=0.5pt,arc=.2em,boxsep=1pt,left=2pt,right=2pt,top=2pt,bottom=2pt]

\begin{scriptsize}
\begin{Verbatim}[commandchars=\\\{\}]
\PYG{k+kn}{import} \PYG{n+nn}{numpy} \PYG{k+kn}{as} \PYG{n+nn}{np}

\PYG{c+c1}{\PYGZsh{} A 2D point class (x, y)}
\PYG{k}{class} \PYG{n+nc}{PyPoint2D}\PYG{p}{(}\PYG{n+nb}{object}\PYG{p}{):}
    \PYG{c+c1}{\PYGZsh{} constructor}
    \PYG{k}{def} \PYG{n+nf}{\PYGZus{}\PYGZus{}init\PYGZus{}\PYGZus{}}\PYG{p}{(}\PYG{n+nb+bp}{self}\PYG{p}{,} \PYG{n}{x}\PYG{p}{,} \PYG{n}{y}\PYG{p}{):}
        \PYG{n+nb+bp}{self}\PYG{o}{.}\PYG{n}{x} \PYG{o}{=} \PYG{n+nb}{float}\PYG{p}{(}\PYG{n}{x}\PYG{p}{)}
        \PYG{n+nb+bp}{self}\PYG{o}{.}\PYG{n}{y} \PYG{o}{=} \PYG{n+nb}{float}\PYG{p}{(}\PYG{n}{y}\PYG{p}{)}
    \PYG{c+c1}{\PYGZsh{} local coordinate dimension}
    \PYG{k}{def} \PYG{n+nf}{dim}\PYG{p}{(}\PYG{n+nb+bp}{self}\PYG{p}{):}
        \PYG{k}{return} \PYG{l+m+mi}{2}
    \PYG{c+c1}{\PYGZsh{} map manifold point other to local coordinate}
    \PYG{k}{def} \PYG{n+nf}{local}\PYG{p}{(}\PYG{n+nb+bp}{self}\PYG{p}{,} \PYG{n}{other}\PYG{p}{):}
        \PYG{k}{return} \PYG{n}{np}\PYG{o}{.}\PYG{n}{array}\PYG{p}{([}\PYG{n}{other}\PYG{o}{.}\PYG{n}{x} \PYG{o}{\PYGZhy{}} \PYG{n+nb+bp}{self}\PYG{o}{.}\PYG{n}{x}\PYG{p}{,} \PYG{n}{other}\PYG{o}{.}\PYG{n}{y} \PYG{o}{\PYGZhy{}} \PYG{n+nb+bp}{self}\PYG{o}{.}\PYG{n}{y}\PYG{p}{],} \PYG{n}{dtype}\PYG{o}{=}\PYG{n}{np}\PYG{o}{.}\PYG{n}{float}\PYG{p}{)}
    \PYG{c+c1}{\PYGZsh{} apply changes in local coordinate to manifold, \PYGZbs{}oplus operator}
    \PYG{k}{def} \PYG{n+nf}{retract}\PYG{p}{(}\PYG{n+nb+bp}{self}\PYG{p}{,} \PYG{n}{vec}\PYG{p}{):}
        \PYG{k}{return} \PYG{n}{PyPoint2D}\PYG{p}{(}\PYG{n+nb+bp}{self}\PYG{o}{.}\PYG{n}{x} \PYG{o}{+} \PYG{n}{vec}\PYG{p}{[}\PYG{l+m+mi}{0}\PYG{p}{],} \PYG{n+nb+bp}{self}\PYG{o}{.} \PYG{o}{+} \PYG{n}{vec}\PYG{p}{[}\PYG{l+m+mi}{1}\PYG{p}{])}
\end{Verbatim}
\end{scriptsize}

\end{tcolorbox}
\end{figure*}

\end{document}